# Motion Control of Interactive Robotic Arms Based on Mixed Reality Development

Hanxiao Chen
*Department of Automation*
*Harbin Institute of Technology*
Harbin, China
hanxiaochen@hit.edu.cn

*Abstract*—Mixed Reality(MR) is constantly evolving to inspire new patterns of robot manipulation for more advanced Human-Robot Interaction under the 4[th] Industrial Revolution Paradigm. Consider that Mixed Reality aims to connect physical and digital worlds to provide special immersive experiences, it is necessary to establish the information exchange platform and robot control systems within the developed MR scenarios. In this work, we mainly present multiple effective motion control methods applied on different interactive robotic arms (e.g., UR5, UR5e, myCobot) for the Unity-based development of MR applications, including GUI control panel, text input control panel, end-effector object dynamic tracking and ROS-Unity digital-twin connection.

*Keywords—Mixed Reality, Motion Control, Robotic Arm*

## I. Introduction

Industrial robots have presented considerable applications for diverse fields consisting of smart manufacturing and space exploration. Moreover, robotic manipulation has been explored comprehensively based on new technology advances including deep reinforcement learning [1, 2, 3], human-robot interaction [4, 5] and extended reality [6, 7, 8, 9]. To emphasize, motion control of robot arms stands as the most significant section for robotic manipulation since it means the software component of a robotic system which dictates how a robot can move through the actions of rotating and sliding joints to finish missions. Therefore, multiple prominent motion control approaches have emerged from inverse kinematics [10], motion planning [11], imitation learning and reinforcement learning.

Based on the magical function of blending both virtual and physical worlds, Mixed Reality has harvested a lot of attention to empower the industry with more advanced human-robot interaction (HRI). The market is exploding with more MR head-mounted devices and apps, even [12] has demonstrated the state-of-the-art applications and future prospects of the Microsoft HoloLens 2 device in medical and healthcare context. Alongside such expanding technology, digital twin is proposed to offer the digital representation of a physical environment, object, process, or service that behaves and looks like its counterpart in the real-world. Moreover, [7] presents a digital twin approach applied successfully with Microsoft Hololens 2 and KUKA IIWA for efficient human-robot interaction. To tackle the issue of flawless and safe manipulation of robotic arms within real-time collaborative environments, [8] indicates closed-loop robotic arm manipulation based on mixed reality and ROS with the UR10 robot case study. [9] even introduces a framework for interactive control of multiple robots with industrial manipulators, mobile robots, and unmanned aerial vehicles (UAV) by using Robotic Operation System (ROS) and Microsoft HoloLens device.

To facilitate the MR-based robotic applications, we aim to focus on motion control of robotic arms to provide effective and efficient control for applicable human-robot interaction based on the Unity 3D MR development scenes, which can be transferred to multiple types of robots (Fig. 1). Inspired by the fundamental robot control solutions [10, 11], we develop two mainstream methodologies including "Control by Unity" and "Control by ROS" to integrate it with the MRTK-Unity toolkit smoothly, even achieve the successful end-effector dynamic object tracking and ROS-Unity digital twin interactions via diverse MR scenarios, which can lay a good foundation for the development of complicated industrial manipulation interfaces.

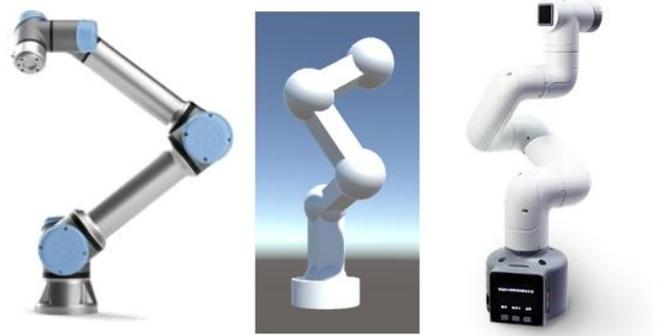

Fig. 1. Interactive robotic arms: UR5e, self-built arm, myCobot.

Our paper is structured as follows. In Section II, the most pertinent literature of robot motion control and our proposed two integration frameworks for MR development are discussed in detail. Section III presents the experimental results with different interactive robotic arms to validate the effectiveness of our proposed motion control pipelines in Unity-based MR scenarios. Thus, Section IV summarizes the whole research project and discusses our future work.

## II. Methods

Motion control of robotic arms has developed from the traditional Forward/Inverse Kinematics[10] to complex motion planning [11], and we briefly review such detailed methods as follows and investigate how to take good advantage of them on general robotic arms for interaction in MR scenarios.

*A. Inverse Kinematics*

Defined as a mathematical process, Inverse Kinematics (IK) attempts to find configurations for an end-effector to reach a Cartesian target. At first, we consider the joint angles of a robot as "Joint Space" whereas "Cartesian Space" consists of the orientation matrix and position vector of its end-effector. With the established relationships, IK aims to convert the position and orientation of a manipulator end-effector from "Cartesian Space" to "Joint Space" and there are two approaches namely "geometric" and "algebraic" applied for deriving the inverse kinematics solution analytically [10].

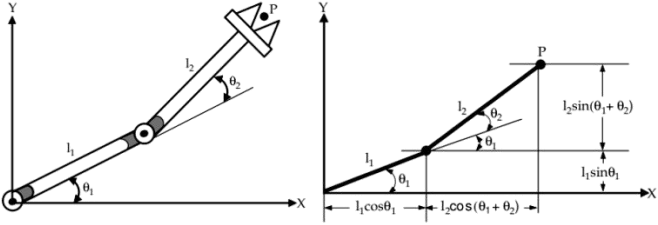

Fig. 2. The geometric IK method explanation [10].

The geometric method is usually applied to the simple 2-DOF planer manipulator (Fig. 2) and decomposes the spatial geometry of the robot manipulator into several plane geometry problems. According to a series of geometric computation, we could obtain the possible solutions for $\theta_1$ and $\theta_2$ to reach the desired position $P(P_x, P_y)$.

$$\theta_1 = A\tan 2\left(\pm\sqrt{1-\left(\frac{p_x(l_1+l_2 c\theta_2)+p_y l_2 s\theta_2}{p_x^2+p_y^2}\right)^2}, \frac{p_x(l_1+l_2 c\theta_2)+p_y l_2 s\theta_2}{p_x^2+p_y^2}\right) \quad (1)$$

$$\theta_2 = A\tan 2\left(\pm\sqrt{1-\left(\frac{p_x^2+p_y^2-l_1^2-l_2^2}{2l_1 l_2}\right)^2}, \frac{p_x^2+p_y^2-l_1^2-l_2^2}{2l_1 l_2}\right) \quad (2)$$

Towards the robot manipulators with complex 3D structure and multiple DOFs, the algebraic approach is used for rigorous derivation, which requires transformation matrix and the robot DH Table for computation. Take the 6-axis robot manipulator (Fig. 3) as example, the position and orientation of the end-effector with respect to the base is given by (3).

$$^0_6T = {^0_1}T(q_1){^1_2}T(q_2){^2_3}T(q_3){^3_4}T(q_4){^4_5}T(q_5){^5_6}T(q_6) \quad (3)$$

To find the inverse kinematics solution for the first joint ($q_1$) as a function of the known elements of $^{base}_{end-effector}T$, the link transformation inverses are premultiplied as follows.

$$[^0_1T(q_1)]^{-1} {^0_6}T = {^1_2}T(q_2){^2_3}T(q_3){^3_4}T(q_4){^4_5}T(q_5){^5_6}T(q_6) \quad (4)$$

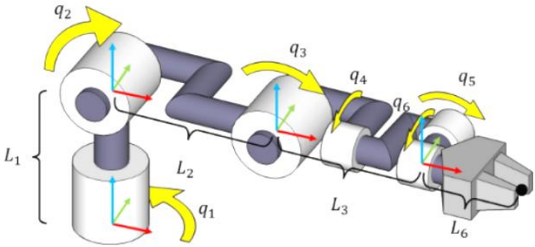

Fig. 3. The algebraic IK method explanation on 6-DOF manipulator.

As for other variables, the following equations are obtained as a similar manner.

$$[^0_1T(q_1){^1_2}T(q_2)]^{-1} {^0_6}T = {^2_3}T(q_3){^3_4}T(q_4){^4_5}T(q_5){^5_6}T(q_6) \quad (5)$$

$$[^0_1T(q_1){^1_2}T(q_2){^2_3}T(q_3)]^{-1} {^0_6}T = {^3_4}T(q_4){^4_5}T(q_5){^5_6}T(q_6) \quad (6)$$

$$[^0_1T(q_1){^1_2}T(q_2){^2_3}T(q_3){^3_4}T(q_4)]^{-1} {^0_6}T = {^4_5}T(q_5){^5_6}T(q_6) \quad (7)$$

$$[^0_1T(q_1){^1_2}T(q_2){^2_3}T(q_3){^3_4}T(q_4){^4_5}T(q_5)]^{-1} {^0_6}T = {^5_6}T(q_6) \quad (8)$$

Then, we get 12 simultaneous set of nonlinear equations to be solved and the 12 nonlinear matrix elements of right hand side are either zero, constant or functions of $q_2$ through $q_6$. If the elements on the left hand side which are the function of $q_1$ are equated with the elements on the right hand side, then the joint variable $q_1$ can be solved as functions of $r_{11}, r_{12}, \ldots r_{33}$, $p_x$, $p_y$, $p_z$ and the fixed link parameters. Once $q_1$ is found, then the other joint variables are solved by the same way as before.

### B. Motion Planning

As demonstrated in [13], motion planning aims to insert a sequence of intermediate points for control between the given path endpoints to achieve smooth movement for robotic arms. In general, motion planning includes the path planning (space) and trajectory planning (time) to produce a collision-free path connecting the initial and goal configurations. Most recently, more motion planning approaches have been proposed and applied within manufacturing. For example, [14] proposes the lower bound tree-RRT (LBT-RRT) as a single-query sampling-based asymptotically near-optimal motion-planning algorithm with high-quality solutions and shorter running time compared with RRT [15]. Moreover, [16] introduces a physics-based motion planning strategy that could provide the collision-free gripping trajectories in uncertain and chaotic environments.

Inspired by reinforcement learning, more plausible motion planning solutions for robotic arms have been developed to improve performance and efficiency, even can be divided into two categories: model-based and model-free motion planning. Based on a 6 DoF sensor-equipped industrial manipulator, [17] presents a novel model-based planning method for industrial inspection applications by addressing view planning and path planning problems simultaneously via a sequencing SCTSP formulation with ROS MoveIt! [18] and the installed Open Motion Planning Library [19] to generate the time-efficient, collision-free motion plans. Regarding the model-free methods, [20] designs a model-free controller to be applied on a 6 DOF PUMA 560 robot to track a desired trajectory effectively with the related intelligent Proportional, Integral and Derivative (iPID) regulators. Moreover, the model-free methods can solve sudden model changes compared with those model-based solutions, but they may take a long time to find a coefficient that satisfies the control task.

Based on the mature methodologies for Inverse Kinematics and Motion Planning, we mainly develop MR applications based on the Unity3D Engine and MRTK-Unity, which is a Microsoft-driven project that provides a set of components and features to accelerate cross-platform MR app development in Unity. To perfectly control the motion of robotic arms via MR scenarios for human-robot interaction, we successfully design and implement the following two motion control frameworks.

*1) Control by Unity:* Since Unity 3D is a powerful game engine with complete graphic interfaces and level-building features, we could design the "Control by Unity" framework (Fig. 4) with two significant "Interactive GUI" and "Inverse Kinematics" modules to achieve successful motion control on robotic arms via the self-built applicable MR development scenarios. Actually, "Interactive GUI" aims to visually present the robot controlling process for interaction whereas "Inverse Kinematics" can be implemented by Unity C# programming scripts for efficient motion control. Therefore, we mainly implement the control section by three effective patterns: (1) GUI Slider Control Panel; (2) Text Input Control Panel; (3) Unity C# Script Control.

Firstly, MRTK-Unity is able to provide users with an immersive operation environment and graphical user interface (GUI) is the most intuitive method for visualization. Consider that robotic arms can be directly controlled by joints for spatial positions, we design the "GUI slider control panel" with the corresponding quantity of robot DoFs to make users easily adjust joint states respectively by sliding each slider. Secondly, we design the "Text Input Control Panel" inspired by the regular text information interaction that inputs the desired position and angular numbers into text boxes for control. After inputting the required information, the robotic arm can be controlled by the "Inverse Kinematics" module to reach the proposed state interactively. Next, for "Unity C# Script Control", we make the sphere present "target" and prepare "Inverse Kinematics" motion control C# scripts for the robot arm to track target positions dynamically so that within MR scenes the robot end-effector can achieve object tracking when the user freely operates the "sphere" target with holographic hand rays and gestures.

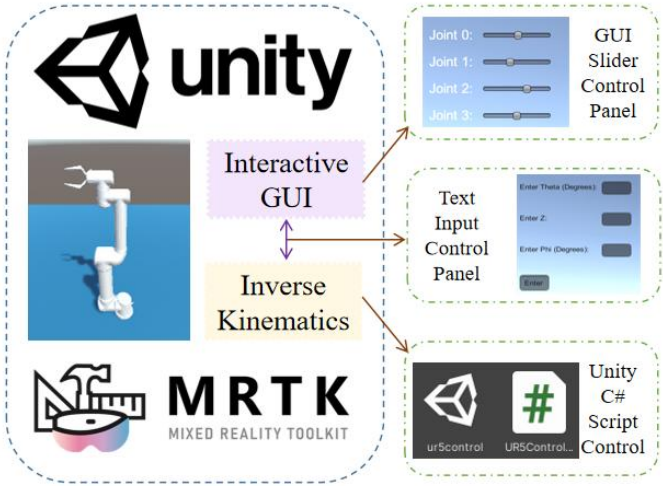

Fig. 4. "Control by Unity" framework.

*2) Control by ROS:* Aside by the Unity, Robot Operating System (ROS) is also extensively applied for robot simulation, hardware abstraction and control system implementation due to its abundant program packages and libraries. Inspired by [7], we design the "Control by ROS" framework according to Digital Twin (DT), which can create the corresponding virtual representation of a real-world physical model using data and information to help users collaboratively design and operate complicated systems in interactive and immersive ways. As illustrated in Fig. 5, we establish the digital-twin connection with Unity Robotics Hub and ROSBridge to match the cross-platform virtual robot models, then apply ROS MoveIt motion planning [18, 19] with Rviz or Python scripts for robot control to simultaneously move the Unity MR-developed robot arm.

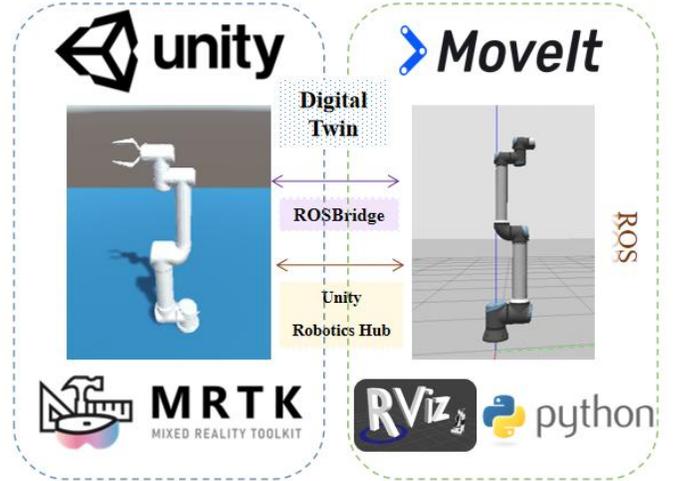

Fig. 5. "Control by ROS" framework.

## III. EXPERIMENTS

To validate the effectiveness of our proposed robotic arm motion control patterns via the MR development, we conduct extensive experiments with diverse interactive robot arms (e.g., UR5, UR5e and myCobot) based on the Windows Unity 2020.3.24f1 platform as follows.

### A. Control by Unity

*1) GUI Slider Control Panel:* As illustrated in Fig. 6, we firstly establish the intuitive graphical user interface (GUI) control panel for the 6 DoF UR5 robot with 6 sliders along to each corresponding joints. While building MR scenarios with

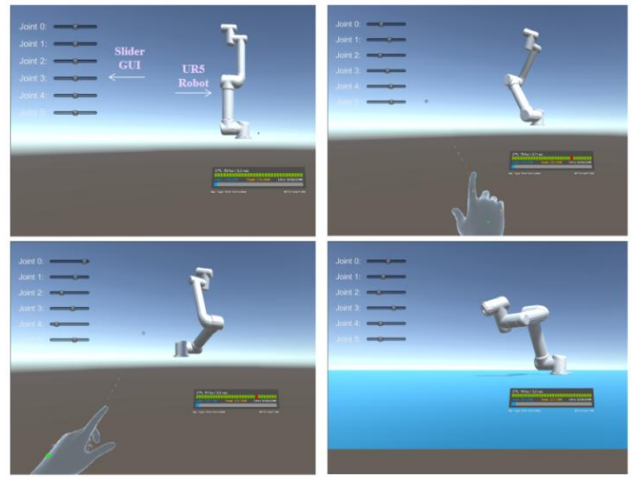

Fig. 6. "GUI Slider Control Panel" experiments.

MRTK-Unity, we create the C# script for motion control to change robot joint states with the visualized GUI sliders. In experiments, users can freely move the sliders to achieve any different UR5 states and view the MR 3D virtual robot model by the similar Microsoft Hololens 2 interaction operations (e.g., hand tracking), which means the GUI Slider Control Panel can effectively and directly control the robot motion.

*2) Text Input Control Panel:* In some cases, we need to quantitatively control the robotic arm via Unity MR scenes instead of the GUI qualitative pattern. Thus, we create the Text Input Control Panel for the UR5 robot within Fig. 7 to make users input 3D position and angle variables in text boxes, then the integrated Unity Inverse Kinematics C# algorithm script will drive the robot towards the desired posture relative to its initial coordinate system. Fig. 7 demonstrates the UR5 initial horizontal state and three different motion control results after users input diverse text commands as shown in Table I, even representing the feasibility and effectiveness of the "Text Input" motion control method with mixed reality interactions.

TABLE I. TEXT INPUT COMMANDS FOR THREE SCENARIOS

| Parameters | X | Psi | Y | Theta | Z | Phi |
|---|---|---|---|---|---|---|
| (a) | 3 | 0 | 4 | 0 | 5 | 0 |
| (b) | 3 | 0 | -4 | 0 | 5 | 90 |
| (c) | -3 | 60 | 4 | 0 | -2 | 0 |

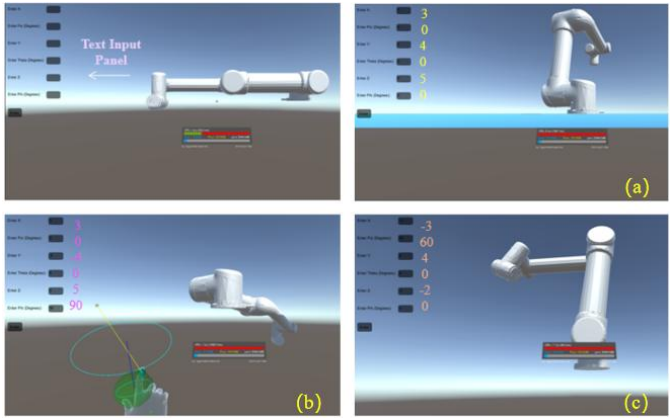

Fig. 7. "Text Input Control Panel" experiments.

*3) Unity C# Script Control:* Within the Unity MR platform, we also self-develop a 4 DoF robot arm and a target red ball (Fig. 8) to successfully achieve the end-effector dynamic object tracking by Unity inverse kinematics control scripts when users freely operate the target ball in MR 3D space. To begin with, the robotic arm and the target ball both stay at initial positions. Then once our built MR scene starts running, the robotic end-effector will quickly turn towards the red target ball while keeping a safe distance. Based on MRTK-Unity, we can touch the target ball by hologram hand ray to move it around space so that the IK Unity C# script can dynamically make the robot arm track this target object with reasonable postures no matter where, which indeed lays a good and solid foundation for more complicated MR robotic applications requiring object tracking.

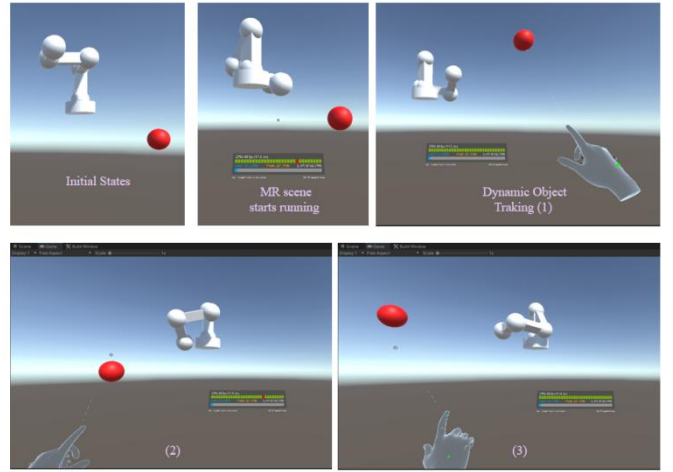

Fig. 8. "Unity C# Script Control" for MR dynamic object tracking.

### B. Control by ROS

Transferred to the designed "Control by ROS" benchmark, we validate the digital-twin method with UR5e and myCobot via the Windows Unity 2020.3.24f1 platform and Oracle VM VirtualBox Ubuntu system installed with ROS. To begin with, we respectively develop two MR environments for myCobot (Fig. 9) and UR5e (Fig. 10), then take advantage of the Unity Robotics Hub and ROSBridge to establish suitable information exchange channels between Unity and ROS for communication.

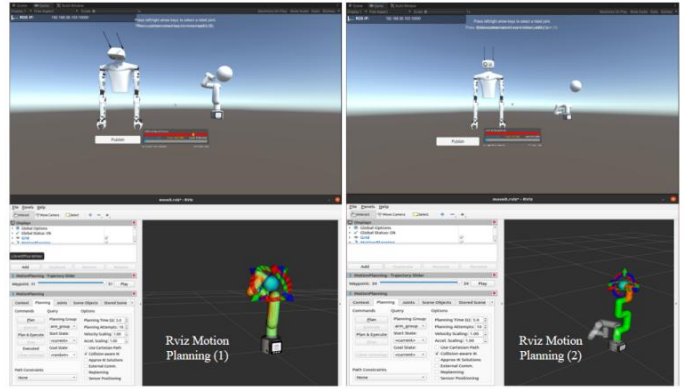

Fig. 9. "Control by ROS" myCobot MR digital-twin scenario.

Obviously, the double arrow aside by "ROS IP" in the upper left corner of Unity interface will appear green once the digital twin connection has been successfully set up. Then, users can launch the ROS MoveIt Rviz tool to conduct visualized motion planning for the ROS virtual myCobot (Fig. 9) while Unity models will also move simultaneously with the same trajectory planned by ROS for target states and positions. Moreover, it is also flexible to achieve precise motion control by executing python scripts via ROS on UR5e robot (Fig. 10). Compared with the previous "Control by Unity" solution, digital twin method is potential to be applied on ROS-driven physical robots for real-world control when users deploy such Unity developed MR scenarios on Microsoft Hololens 2. Thus, ROS can serve as a medium between the real robotic arm and the

Unity 3D virtual model to realize flexible synchronized digital-twin movement.

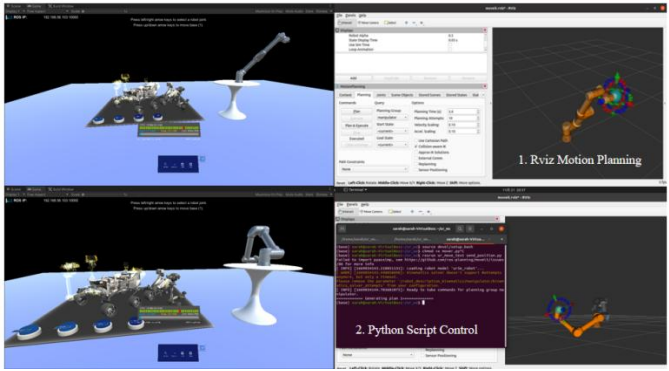

Fig. 10. "Control by ROS" UR5e MR digital-twin scenario.

IV. CONCLUSION

In summary, we propose and successfully implement two different patterns of motion control for interactive robotic arms (e.g., UR5, UR5e, myCobot) via mixed reality development: "Control by Unity" and "Control by ROS". Based on the core modules of "Interactive GUI" and "Inverse Kinematics", our designed "Control by Unity" solutions including "GUI Slider Control Panel", "Text Input Control Panel" and "Unity C# Script Control" can successfully achieve intuitive and effective motion control for UR5 and UR5e robots with interaction, even realize the MR dynamic end-effector object tracking on our self-built robotic arm via MRTK-Unity. Transferred to the digital-twin "Control by ROS" framework, we create complex MR scenes with UR5e and myCobot, even control the Unity robot motion according to ROS Rviz and python scripts. For future work, we will deploy the designed Unity MR scenarios on Microsoft Hololens 2 with proper adjustments and apply the proposed motion control methods on physical robots.

REFERENCES

[1] Nguyen H, La H. Review of deep reinforcement learning for robot manipulation[C]//2019 Third IEEE International Conference on Robotic Computing (IRC). IEEE, 2019: 590-595.
[2] Chen H. Robotic manipulation with reinforcement learning, state representation learning, and imitation learning (student abstract)[C]//Proceedings of the AAAI Conference on Artificial Intelligence. 2021, 35(18): 15769-15770.
[3] Gu S, Holly E, Lillicrap T, et al. Deep reinforcement learning for robotic manipulation with asynchronous off-policy updates[C]//2017 IEEE international conference on robotics and automation (ICRA). IEEE, 2017: 3389-3396.
[4] Goodrich M A, Schultz A C. Human–robot interaction: a survey[J]. Foundations and Trends® in Human–Computer Interaction, 2008, 1(3): 203-275.
[5] Chen H, Wang J, Meng M Q H. Kinova Gemini: Interactive Robot Grasping with Visual Reasoning and Conversational AI[C]//2022 IEEE International Conference on Robotics and Biomimetics (ROBIO). IEEE, 2022: 129-134.
[6] Tu J, Autiosalo J, Ala-Laurinaho R, et al. TwinXR: Method for using digital twin descriptions in industrial eXtended reality applications[J]. Frontiers in Virtual Reality, 2023.
[7] Gallala A, Kumar A A, Hichri B, et al. Digital Twin for human–robot interactions by means of Industry 4.0 . Enabling Technologies[J]. Sensors, 2022, 22(13): 4950.
[8] Mourtzis D, Angelopoulos J, Panopoulos N. Closed-Loop Robotic Arm Manipulation Based on Mixed Reality[J]. Applied Sciences, 2022, 12(6): 2972.
[9] Ostanin M, Yagfarov R, Devitt D, et al. Multi robots interactive control using mixed reality[J]. International Journal of Production Research, 2021, 59(23): 7126-7138.
[10] Kucuk S, Bingul Z. Robot kinematics: Forward and inverse kinematics[M]. London, UK: INTECH Open Access Publisher, 2006.
[11] Li Z, Canny J F, Sastry S S. On motion planning for dexterous manipulation[C]//Proceedings of the IEEE Conference on Robotics and Automation. 1989: 775-780.
[12] Palumbo A. Microsoft Hololens 2 in medical and healthcare context: state of the art and future prospects[J]. Sensors, 2022, 22(20): 7709.
[13] Liu S, Liu P. A review of motion planning algorithms for robotic arm systems[C]//RiTA 2020: Proceedings of the 8th International Conference on Robot Intelligence Technology and Applications. Singapore: Springer Singapore, 2021: 56-66.
[14] Salzman O, Halperin D. Asymptotically near-optimal RRT for fast, high-quality motion planning[J]. IEEE Transactions on Robotics, 2016, 32(3): 473-483.
[15] LaValle S M, Kuffner Jr J J. Randomized kinodynamic planning[J]. The international journal of robotics research, 2001, 20(5): 378-400.
[16] Moll M, Kavraki L, Rosell J. Randomized physics-based motion planning for grasping in cluttered and uncertain environments[J]. IEEE Robotics and Automation Letters, 2017, 3(2): 712-719.
[17] Jing W, Polden J, Tao P Y, et al. Model-based coverage motion planning for industrial 3D shape inspection applications[C]//2017 13th IEEE Conference on Automation Science and Engineering (CASE). IEEE, 2017: 1293-1300.
[18] Chitta S, Sucan I, Cousins S. Moveit![ros topics][J]. IEEE Robotics & Automation Magazine, 2012, 19(1): 18-19.
[19] Sucan I A, Moll M, Kavraki L E. The open motion planning library[J]. IEEE Robotics & Automation Magazine, 2012, 19(4): 72-82.
[20] Abouaïssa H, Chouraqui S. On the control of robot manipulator: A model-free approach[J]. Journal of Computational Science, 2019, 31: 6-16.